\documentclass[10pt,lettersize,journal]{IEEEtran}
\usepackage{amsmath,amsfonts}
\usepackage{algorithmic}
\usepackage{algorithm}
\usepackage{array}
\usepackage[caption=false,font=normalsize,labelfont=sf,textfont=sf]{subfig}
\usepackage{textcomp}
\usepackage{stfloats}
\usepackage{url}
\usepackage{verbatim}
\usepackage{graphicx}
\usepackage[utf8]{inputenc}
\usepackage{amssymb}
\usepackage{amsmath}
\usepackage{adjustbox}
\usepackage{multirow}
\usepackage{bm}
\usepackage{xcolor}
\usepackage{siunitx}
\usepackage{colortbl}
\usepackage{color, colortbl}
\definecolor{Orange}{rgb}{0.0,0.8,0.1}
\definecolor{red}{rgb}{1.0, 0.0, 0.0}
\definecolor{LightOrange}{rgb}{0.8,0.7,0.1}
\definecolor{green}{rgb}{0.0, 1.0, 0.0}

\newcommand{\sao}[1]{{\color{black}#1}} 
\newcommand{\mrh}[1]{{\color{black}#1}} 
\usepackage{cite}
\def\BibTeX{{\rm B\kern-.05em{\sc i\kern-.025em b}\kern-.08em
    T\kern-.1667em\lower.7ex\hbox{E}\kern-.125emX}}
\usepackage{balance}
\begin{document}

\title{ISC-Perception: A Hybrid Computer Vision Dataset for Object Detection in Novel Steel Assembly}

\author{Miftahur~Rahman$^{1}$,
        Samuel~Adebayo$^{1}$~\IEEEmembership{Member,~IEEE,}
        Dorian~A.~Acevedo-Mejia$^{2}$,
        David~Hester$^{1}$,\\
        Daniel McPolin$^{1}$,
        Karen Rafferty$^{3}$,
        and Debra F. Laefer$^{4}$
\IEEEcompsocitemizethanks{\IEEEcompsocthanksitem $^{1}$School of Natural and Built Environment, Queen's University Belfast, Belfast, United Kingdom .\protect\\
E-mail: \{miftahur.rahman, samuel.adebayo, d.hester, d.mcpolin\}@qub.ac.uk
\IEEEcompsocthanksitem $^{2}$School of Civil \& Environmental Engineering, and Construction Management, University of Texas at San Antonio, USA. \protect\\
Email: dorian.acevedo@utsa.edu
\IEEEcompsocthanksitem $^{3}$School of Electronics, Electrical Engineering, and Computer Science, Queen's University Belfast, Belfast, United Kingdom.\protect\\
Email: k.rafferty@qub.ac.uk
\IEEEcompsocthanksitem $^{4}$Department of Civil and Urban Engineering, New York University, USA.\protect\\
Email: debra.laefer@nyu.edu\protect\\
}

}

\maketitle

\begin{abstract}
The Intermeshed Steel Connection (ISC) system, when paired with robotic manipulators, can accelerate steel-frame assembly and improve worker safety by eliminating manual assembly. Dependable perception is one of the initial stages for ISC-aware robots. However, this is hampered by the absence of a dedicated image corpus, as collecting photographs on active construction sites is logistically difficult and raises safety and privacy concerns. In response, we introduce ISC-Perception, the first hybrid dataset expressly designed for ISC component detection. It blends procedurally rendered CAD images, game-engine photorealistic scenes, and a limited, curated set of real photographs, enabling fully automatic labelling of the synthetic portion. \sao{We explicitly account for all human effort to produce dataset, including simulation engine and scenes setup, asset preparation, post-processing scripts and quality checks; our total human time to generate a 10{,}000-image dataset was 30.5\,h versus 166.7\,h for manual labelling at 60\,s per image (-81.7\%). A manual pilot on a representative image with five instances of ISC members took 60\,s (maximum 80\,s), anchoring the manual baseline.}. Detectors trained on ISC-Perception achieved a mean Average Precision at IoU 0.50 of 0.756, substantially surpassing models trained on synthetic-only or photorealistic-only data. \sao{On a 1{,}200-frame bench test, we report mAP@0.50/mAP@[0.50:0.95] of 0.943/0.823 }. By bridging the data gap for construction-robotics perception, ISC-Perception facilitates rapid development of custom object detectors and is freely available for research and industrial use upon request.
\end{abstract}


\section{Introduction}\label{sec:introduction}
\subsection{Background and Motivation}\label{subsec:trad-steel-building}
\IEEEPARstart{R}{obotic} manipulators have transformed factory‐based manufacturing, yet their impact on construction remains modest. Unlike shop floors, building sites are unstructured, weather‑exposed, and governed by stringent safety constraints, all of which complicate autonomous operation. Steel frame erection, which is one of the most labour‑intensive, high‑risk phases of a build, stands to benefit most from automation: cranes dominate the critical path while transporting heavy steel items, bolting demands skilled crews working at height, and schedule delays cascade to all downstream trades \cite{liang_ras_2017}. A robot capable of recognising, grasping, and interlocking structural members in situ could shorten crane time, lower accident rates, and mitigate skilled‑labour shortages.

In building construction, structural steel frames are composed of individual beams and columns that are typically assembled on-site due to the challenges associated with transporting large assemblies. This process involves several key steps: (1) identifying and lifting each structural element from the storage area, (2) transporting it to the installation location; (3) aligning it with the existing structure, and (4) fastening it to the structural frame using bolts or welds \cite{liang_ras_2017}. Although both methods are common for connecting steel members, bolts are generally preferred on-site due to their ease of installation, faster connection times, better quality control, and reduced inspection requirements. However, the extensive use of bolts in structural steel connections introduces additional challenges for the deployment of robots in the field.

\begin{figure*}[!ht]
    \centering
    \includegraphics[width=5.3in]{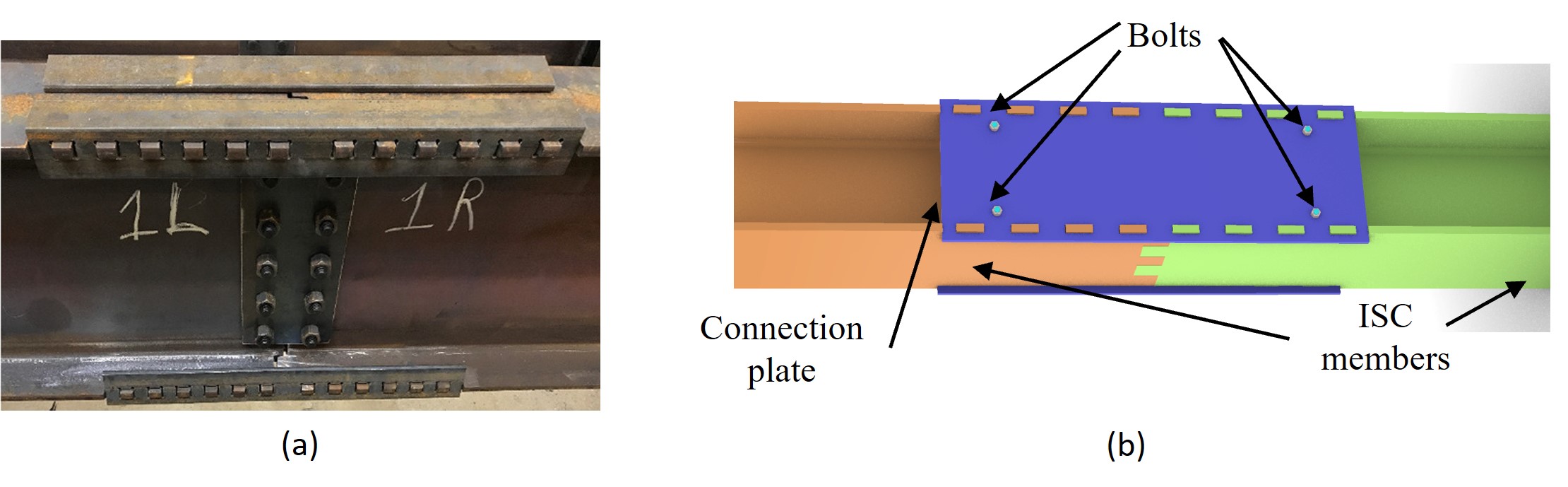}%
    \caption{Components of ISC beam-to-beam; (a) earlier version of fabricated ISC \protect\cite{al-sabah_introduction_2020}, (b) CAD drawing of ISC with single connection.}
    \label{fig:ISC_beam_to_beam}
\end{figure*}

The recently proposed Intermeshed Steel Connection (ISC) system eliminates most of the temporary bolts required by conventional moment or shear splices. The ISC can be manufactured using cutting-edge technologies such as high-density plasma cutting, water jet cutting, and laser cutting \cite{shemshadian_experimental_2020}. ISC has two types of components: ISC member and ISC connection plates. Initial design of ISC has 3 connection plates on one side (Fig. \ref{fig:ISC_beam_to_beam}(a)) but the newer version requires only one connection plates on each side with fewer number of bolts (Fig. \ref{fig:ISC_beam_to_beam}(b)). Precision‑cut male–female tabs guide members into alignment so that only a handful of set‑bolts are needed to secure the joint \cite{al-sabah_introduction_2020,shemshadian_experimental_2020}. By trimming cycle times and tolerating direct reuse, ISC reduces material waste and greenhouse‑gas emissions while preserving structural capacity.  These benefits align with industry trends towards design‑for‑manufacture‑and‑assembly (DfMA) \cite{montazeri_design_2024} and circular construction. However, the ISC's unconventional geometry poses fresh challenges for computer vision: the connection plates are compact, partially occluded once mated, and often coated with reflective galvanisation.

Robots cannot exploit ISC unless they can reliably identify connection plates, member ends, and mating features under dynamic lighting and clutter. Conventional weld or bolt heads provide distinctive geometry; ISC plates, by contrast, are largely planar and differ only by tab pattern. No public image corpus captures these subtleties, and collecting site photographs is fraught with \textit{access restrictions, privacy regulations, and weather‑dependent scheduling}. Consequently, vision models trained on generic construction datasets (e.g., MOCS \cite{noauthor_dataset_nodate}, SODA\cite{noauthor_soda_nodate}) fail to generalise for robotic assembly tasks. Bridging this domain gap requires a \emph{task‑specific dataset} that mixes real jobsite context with photorealistic renders and procedurally generated scenes to achieve both scale and fidelity.

The economic stakes are high: the global structural‑steel market exceeded USD 110B in 2023 and is projected to grow at approximately 5\%, CAGR through 2030, driven by urban densification and industrial expansion~\cite{precedenceresearch_structuralsteel_2025}. While connection labour alone can account for up to 25\% of erection cost~\cite{carter_practical_2004, chambers_connecting_2022}. \mrh{Hence, A vision-enabled robotic ISC assembly workflow has been proposed, which holds the promise of safer jobsites and substantially reduce cost and scedule savings ~\cite{ISC_proposed_framework}}.

\subsection{Images as Fuel for Training Vision Models} \label{subsec:importance-image-for-cv}

Progress in deep learning has been driven less by algorithmic novelty than by the availability of vast, well-annotated image corpora. Construction-robotics vision imposes even steeper data demands: detectors must recognise partially occluded objects, track them under harsh illumination, and generalise across projects that vary by geometry, finish, and weather \cite{jiang_computer_2023,li_application_2020,noauthor_safety_nodate,guo_computer_2021,seo_computer_2015,teizer_right-time_2016}.

Image data plays a pivotal role by forming a foundation for training and testing computer vision and machine learning algorithms, providing essential visual information for tasks such as object detection, image classification, and semantic segmentation \cite{zhong_mapping_2019}. Without access to high-quality and diverse image datasets, the performance (accuracy and generalization) of these algorithms can be significantly hindered. As computer vision technology rapidly advances in the construction industry, the demand for accurate and comprehensive interpretation of construction site imagery has become increasingly urgent \cite{soltani_automated_2016}.

Generic datasets such as COCO or ImageNet misrepresent site reality: they lack steel members, cranes, PPE, and the dense clutter typical of erection yards. Direct transfer can depress mean Average Precision (mAP) by up to 40\% when models are tested on construction imagery \cite{soltani_automated_2016}. Building an in-domain corpus is equally fraught. Cameras are often barred by safety briefings, union rules, or privacy regulations; outdoor shoots hinge on weather windows; and pixel-accurate annotation of high-resolution frames can consume weeks of person-hours \cite{mostafa_review_2021}. The hurdle is steeper still for bespoke components such as the ISC, for which no archival photographs yet exist and whose galvanised surfaces frustrate automated labelling.

Data, not algorithms, has thus become the principal bottleneck. An effective remedy must supply (i) \textit{scale} for deep networks, (ii) \textit{fidelity} to capture ISC’s subtle tab geometry, and (iii) \textit{diversity} in backgrounds, lighting, and occlusions, while curbing manual annotation cost. Section~\ref{subsec:synthetic-image-dataset-shortcomings} surveys how synthetic and photorealistic imagery can satisfy those requirements and where current approaches fall short.

\subsection{Synthetic and Photorealistic Data: Benefits and Pitfalls} \label{subsec:synthetic-image-dataset-shortcomings}
A practical solution to address the challenges of limited access and varying construction site conditions is the creation of annotated synthetic image datasets to supplement real ones \cite{soltani_automated_2016}. These synthetic datasets can be generated using computer graphics techniques, 3D modelling software or game engines enabling the simulation of diverse construction environments with different objects and backgrounds.

Computer vision models trained solely on synthetic images often perform worse than those trained on real images. For example, grocery item detection models trained on 400,000 synthetic images performed less effectively than models trained with only 760 real images \cite{noauthor_210704259_2021}. Yet, combining just 76 real images with the synthetic images produced superior results compared to both models. \mrh{Moreover, randomization techniques (such as lighting condition, weather condition, timing of the day, textures, camera perspective etc.) are used for generating synthetic images which reduce the sim2real gap and improve the diversity of the dataset \mbox{\cite{remmas_PCGOD} \cite{WANG2024123489}}}. Therefore, a hybrid dataset that integrates real and synthetic images could be an effective approach for training computer vision models for construction applications \cite{bayraktar_hybrid_2019}. However, obtaining sufficient real images for many construction scenarios or custom objects, such as the ISC, remains difficult. In such cases, computer-aided design tools can generate and render photorealistic models of custom objects in various settings, reducing the reliance on real images.

Additionally, ISC plates pose an additional hurdle: their galvanised coating creates specular highlights that shift with sun angle, and the laser-cut tab patterns differ by millimetres. Capturing these cues demands high-dynamic-range rendering plus fine surface normal maps—costly to generate at scale. Conversely, photographing ISC plates on active sites remains impractical, because the system is not yet widely deployed. Hence a \emph{hybrid} strategy [(i) auto-generates large volumes of domain-randomised synthetic frames, (ii) injects photorealistic ray-traced scenes for material fidelity, and (iii) enhances the mix with a small, curated set of real photographs] offers the best trade-off between cost and realism.

\vspace{4pt}
\noindent\textbf{Research gap.}  
To date, no public dataset combines these three modalities for steel-connection detection; existing construction corpora (MOCS \cite{noauthor_dataset_nodate}, SODA\cite{noauthor_soda_nodate}) neither model bespoke joints nor provide labels. Bridging this gap is therefore prerequisite to closing the perception loop for robotic ISC assembly.

\subsection{Research Contribution}\label{subsec:research-contribution}
Existing vision datasets in construction focus on equipment or personnel safety. None address robotic assembly of structural steel components such as beams, columns, or ISC plates. We fill this void by devising and releasing \emph{ISC-Perception}, a task-specific, hybrid corpus for object detection in robotic steel erection.

In summary, the main contributions of this paper are:

\begin{itemize}
    \item \sao{A methodology for creating a hybrid dataset for ISC components using real, photorealistic, and synthetic images to tackle the scarcity of real images tailored for robotic assembly tasks, reducing human effort from 166h (manual 10,000 images at 60s per image) to 30.5h with our Unity-based pipeline (~81.7\%); see Table~\ref{tab:time_to_dataset}}
    \item The analysis of training performance of computer vision algorithms for different types of images and validation of the trained computer vision model in small-scale setup.
\end{itemize}

To contextualise these contributions, the next section \ref{sec:background-of-the-paper}, reviews the current state of the art in real and synthetic computer vision datasets. Subsequently, the section \ref{sec:hybrid-methods} discusses the procedural approach for generating the hybrid dataset. Section \ref{sec:Dataset} provides insight on the ISC-Perception dataset. Section \ref{sec:Performance-Analysis} reviews the outcomes of the training and testing phases, followed by a discussion of the results and findings. Finally, Section \ref{sec:conclusions} of the paper summarizes the research results and their significant impacts on the construction industry.

\section{Literature Review}
\label{sec:background-of-the-paper}
This section provides an overview of prominent general purpose computer vision datasets (see \ref{subsec:CV-dataset}) and datasets specific to construction industry (see \ref{subsec:CV-in-construction}). 

\subsection{Computer Vision Datasets}\label{subsec:CV-dataset}
As this research focuses on generating an image dataset for ISC, this section provides a comprehensive overview of prominent computer vision datasets. Computer vision datasets can be broadly categorized into two main types: real-world datasets and synthetic datasets.

Real-world datasets consist of images captured from actual environments and are crucial for training and evaluating models across a range of tasks, including object recognition, object detection, segmentation, and scene understanding \cite{russakovsky_imagenet_2015}. Numerous widely used datasets have been developed to support these tasks. Prominent examples include ImageNet \cite{deng_imagenet_2009}, COCO (Common Objects in Context) \cite{lin_microsoft_2014}, Pascal Visual Object Classes \cite{everingham_pascal_2010}, Open Images \cite{kuznetsova_open_2020}, Cityscapes \cite{cordts_cityscapes_2016}, and KITTI \cite{geiger_vision_2013}. Table~\ref{tab:dataset_overview} provides an overview of these key datasets, highlighting their specific features and contributions to the field.

\begin{table*}
\caption{Overview of Popular Datasets for Computer Vision Tasks}
\label{tab:dataset_overview}
\centering
\begin{tabular}{|l|l|c|c|l|l|l|}
\hline
\textbf{Dataset} & \textbf{Purpose} & \textbf{Year} & \textbf{Classes} & \textbf{Images} & \textbf{Annotations} & \textbf{Domain} \\ 
\hline
ImageNet & Object recognition/classification & 2009 & 21,841 & 14,197,122 & Bounding boxes & General \\
COCO & Object detection/segmentation & 2014 & 80 & 328,000 & Bboxes, masks & General \\
Pascal VOC & Object detection/classification & 2005 & 20 & 11,530 & Bounding boxes & General \\
Open Images & Object detection/classification & 2016 & 19,958 & 9,011,219 & Bounding boxes & General \\
KITTI & Autonomous driving & 2012 & 9 & 7,481 & 3D/2D Bounding Boxes & Urban driving \\
Cityscapes & Semantic segmentation & 2016 & 30 & 5,000 & Segmentation masks & Urban \\
\hline
\end{tabular}
\end{table*}

Synthetic dataset generation in computer vision involves creating artificial images and annotations using tools such as rendering engines (e.g., Blender \cite{blender}, Unity 3D \cite{unity_perception}, Nvidia Omniverse \cite{nvidia_omniverse}, Unreal Engine \cite{unreal_engine}), physics-based simulation software (e.g., Gazebo \cite{gazebo}, Webots \cite{webots}, CoppeliaSim \cite{coppeliaSim}, and generative AI like GANs. These datasets are particularly valuable for generating large-scale, cost-effective, and safe alternatives to real-world data collection. Examples include synthetic datasets derived from video games like Half-Life 2 \cite{taylor_half-life2}, the SYNTHIA dataset for semantic segmentation \cite{ros_synthia_2016}, Hattori et al.’s 3D pedestrian models using Autodesk 3DS Max \cite{hattori_learning_2015}, and the Virtual-KITTI \cite{gaidon_virtualworlds_2016} and Virtual-KITTI 2 \cite{cabon_virtual_2020} datasets, which replicate urban driving scenes with automated annotations via Unity.

Synthetic datasets can be generated using various 3D CAD model rendering and visualization software, incorporating appropriate lighting and scene generation techniques. For example, Aubry et al. developed a dataset of 86,366 synthesized images by rendering each of 1,393 high-quality 3D chair models from 62 distinct viewpoints \cite{aubry_seeing_2014}. Additionally, Peng et al. explored the influence of pose, colour, textures, and background by training a deep convolutional neural network (CNN) using crowd-sourced 3D CAD models, highlighting the potential of synthetic data in improving model performance \cite{peng_learning_2015}.

\subsection{Computer Vision Dataset in Construction Industry}
\label{subsec:CV-in-construction}
Vision technology has garnered significant interest across multiple sectors, including construction, where its application is transforming how visual data from construction sites is acquired and interpreted. This technology enables the extraction of valuable information such as progress monitoring, object detection, safety condition analysis, and quality control. Through the automated detection and tracking of workers, excavators, cranes, dump trucks, and other equipment, it is possible to efficiently identify unsafe conditions on construction sites \cite{du_hard_2011}, \cite{rezazadeh_azar_automated_2012}, \cite{chi_automated_2011}, \cite{park_construction_2012}.

SODA \cite{noauthor_soda_nodate}, tailored for construction sites, contains 19,846 images of 15 object classes. The Moving Objects in Construction Sites (MOCS) dataset contains 41,668 images depicting 13 types of moving objects, including equipment and workers, commonly found on construction sites \cite{noauthor_dataset_nodate}. Those images were captured using a camera, UAV, and smartphone from 174 different construction sites of dam, bridge, building, tunnel and highway \cite{noauthor_dataset_nodate}. Del et. al created a small dataset of 1048 images comprising 08 different object classes for detecting construction equipment and human \cite{del_savio_dataset_2022}. All these datasets were collected from real construction sites, carefully chosen and edited to remove any privacy information, and manually annotated, which is laborious and time consuming. In contrast, Barrera-Animas and Delgado proposed a method to generate synthetic data sets that closely resembles real-world conditions, using 3D models of construction machinery, workers, site environments and assets, combined with realistic lighting conditions in different seasons \cite{barrera-animas_generating_2023}. However, all of the mentioned datasets focus primarily on detecting construction equipment and workers to ensure safe operations, with none designed specifically for robotic assembly tasks.

\section{Method of Generating Computer Vision Hybrid Dataset}
\label{sec:hybrid-methods}
This research aims to develop a dataset specifically for the robotic assembly of steel structures using ISC. 
A hybrid dataset will be created containing photorealistic images of ISC components, synthetic images from the simulation \sao{environment, and few real images and trained with the YOLOv8 algorithm for a robotic assembly application. }

\subsection{Dataset Composition}
\label{subsec:dataset-composition}

The creation of a robust computer vision dataset often begins with the selection of target objects for detection or segmentation. In this work, the dataset focuses on three main object classes: a) ISC member, b) ISC connection plate, and c) human; as illustrated in Fig. \ref{fig:ISC_beam_to_beam}, the selection of these classes is driven by the requirements of future robotic assembly tasks. We envisage that the robot must accurately identify ISC components for assembly and detect humans to ensure safety compliance. While typical steel construction sites include equipment such as tower cranes, forklifts, and scaffolds, these are excluded from the dataset since the robot will not interact with them.

Given the novelty of ISC, real images of these components are limited in availability. Hence, to address this, the ISC-Perception dataset integrates three types of images from diverse sources: 

\begin{enumerate}
    \item Type 1: Photorealistic images from SolidWorks (SW) Visualize (category 1 or C1)
    \item Type 2: Synthetic images from Unity 
        \begin{itemize}
            \item Built-in randomizers (category 2 or C2)
            \item Custom randomizers (category 3 or C3)
        \end{itemize}
    \item Type 3: Real images 
        \begin{itemize}
            \item From previous project (category 4 or C4) 
            \item Human images from Roboflow Universe Public Dataset (category 5 or C5)
        \end{itemize}
\end{enumerate}

\subsection{Image Generation}
\label{subsec:image-gen}
\begin{figure*}[!ht]
    \centering
    \includegraphics[width=7.in]{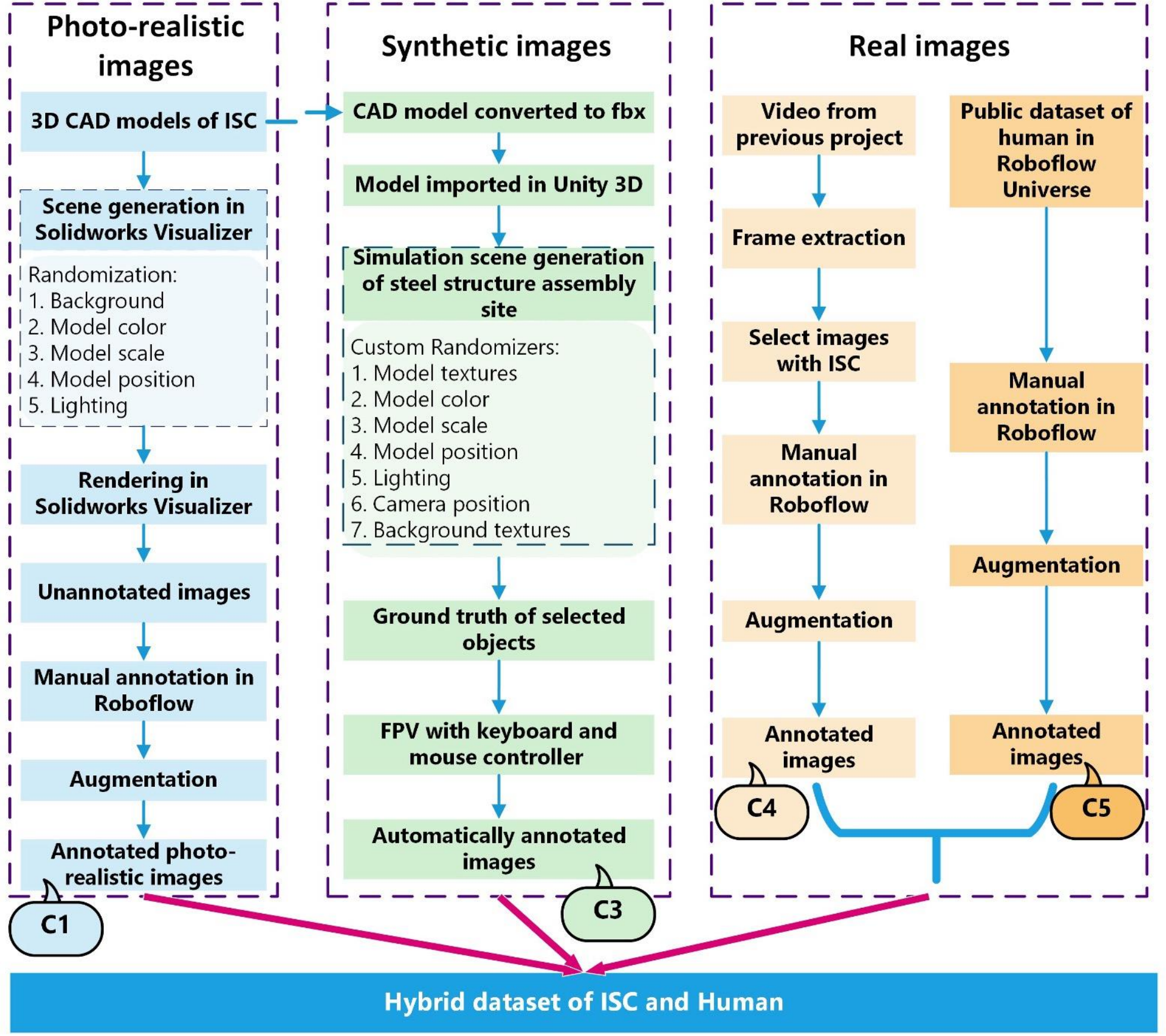}%
    \caption{Source of images and workflow for creating the hybrid dataset combining different types of images.}
    \label{fig:hybrid_dataset_workflow}
\end{figure*}

The image generation workflow is shown in Fig.\ref{fig:hybrid_dataset_workflow}. Synthetic images in Unity (C2) sometimes suffer from jittering, motion blur, and unrealistic appearances (see Supplementary Fig.~S1). To overcome these limitations, custom randomizers (C3) were employed to generate images with enhanced variability across indoor and outdoor assembly scenes. Similarly, photorealistic images (C1) were generated with 3D CAD models in SolidWorks Visualize, incorporating diverse lighting and backgrounds. Finally, the dataset includes manually annotated real images of ISC components and humans, augmented through preprocessing techniques to bolster diversity. This hybrid composition ensures the dataset is both diverse and generalisable, and provides real-world authenticity to support the development of vision systems capable of detecting ISC components in complex assembly environments.

\subsubsection{Photorealistic Images from SolidWorks Visualize}\label{subsub:photo-realistic-images}
As previously established, real images of ISC components are scarce, hence we use CAD rendering software enabled by \sao{SolidWorks Visualize} to generate high-quality photorealistic images to supplement the limited availability of real-world ISC data in ISC-Perception. 

The first stage involves the creation of several 3D models of the two main ISC components using CAD software, as shown in Fig. \ref{fig:hybrid_dataset_workflow}. This is then followed by importing the models  into SolidWorks \mrh{Visualize} for scene generation. During this stage, SW \mrh{Visualize} provides extensive randomization options to enhance dataset diversity. For randomization, SW Visualize pick from 9 total backgrounds and 3 model textures (1 metallic texture and 2 featuring rust), rotates between 0 to 360\textdegree, and varies the lighting conditions, see Table~\ref{tab:sw_randomization}. The output images from the Scene Generation stage are then annotated in Roboflow to get ground truth bounding boxes of ISC objects. Finally, the \mrh{photorealistic} images are augmented in Roboflow to add more variations to the dataset.

\begin{table}[htbp]
\caption{Summary of Randomization Options Used in SolidWorks Visualize}
\label{tab:sw_randomization}
\centering
\setlength{\tabcolsep}{3pt}
\begin{tabular}{p{3cm}p{2cm}p{1cm}p{2cm}}
\hline
\textbf{Background} & \textbf{Model Texture} & \textbf{Rotation} & \textbf{Lighting} \\ 
\hline
Nine options: & Three textures: & 0\textdegree--360\textdegree & Two options: \\
$\bullet$ Black background & $\bullet$ Cast carbon steel (red) & & $\bullet$ Day \\
$\bullet$ Empty outdoor parking & $\bullet$ Metal rust 1 & & $\bullet$ Night \\
$\bullet$ Swiss snow & $\bullet$ Metal rust 2 & & \\
$\bullet$ Steel building site & & & \\
$\bullet$ Black/white background & & & \\
$\bullet$ Industrial lot (night) & & & \\
$\bullet$ Inside glass building & & & \\
$\bullet$ Empty indoor garage & & & \\
$\bullet$ Boiler room & & & \\
\hline
\end{tabular}
\end{table}

\subsubsection{Synthetic Images from Unity}\label{subsub:synthetic-images-unity}
While the use of SW \mrh{Visualize} produces high-quality photorealistic images, the process of manual annotation is time-consuming. Unity with its Perception package, offers an efficient alternative for generating large volumes of automatically annotated synthetic images. C3 in Fig.\ref{fig:hybrid_dataset_workflow}, shows the synthetic image generation process. 

\begin{figure*}[!ht]
    \centering
    \includegraphics[width=7.3in]{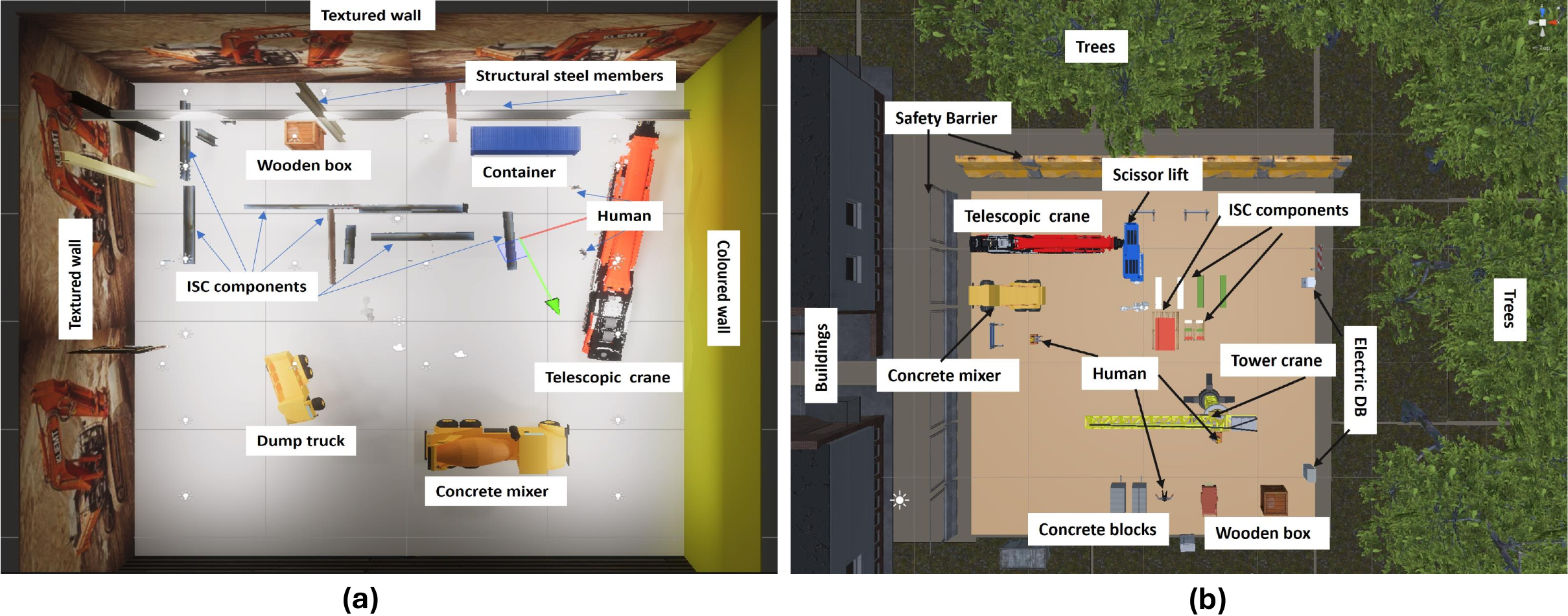}%
    \caption{View of Robotic Steel Assembly in Unity; (a) Outdoor Scene; (b) Indoor Scene}
    \label{fig:robotic-steel-assembly}
\end{figure*}

\textbf{Scene Simulation Generation}: To start with, we used the models built from the CAD software during the generation of photorealistic images and imported those to Unity. Two steel structure assembly simulation scenes were created; one indoor and one outdoor (see Fig. \ref{fig:robotic-steel-assembly} for sample views of robotic steel assembly). The indoor scene included a large workspace with walls displaying custom images to simulate construction environments. Fifty random construction site images were used as wall textures, changing every second to increase variation (Fig.~\ref{fig:robotic-steel-assembly}). The scene was populated with objects such as concrete mixers, dump trucks, scaffolds, and ISC components, placed in various orientations to enhance generalisation. Lighting conditions included directional light mimicking sunlight, dynamically adjusted between -100\textdegree and 100\textdegree, and multiple indoor light sources for a realistic indoor setup.

The outdoor scene contained environmental elements such as trees and buildings alongside construction equipment, safety barriers, and ISC components placed on pallets or the ground. A directional light simulated sunlight, rotating to create shadows from objects like trees and buildings. ISC components were coloured with solid green, red, and white finishes to further diversify the dataset.

The use of custom randomizers addressed the limitations of Unity's built-in randomizers, which failed to generate realistic ISC environments. ISC components were rotated incrementally by 5\textdegree, and background objects were randomly rotated and translated using custom scripts. These randomisations ensured variability in the dataset. Ground truth labels were assigned using Unity’s Perception package. Annotated images were recorded with a first-person camera capturing the scene through a keyboard and mouse-controlled player. 

This approach efficiently produced a diverse dataset of annotated synthetic images, complementing the \mrh{photorealistic} and real images, while addressing the limitations of Unity’s built-in randomizers in simulating real-world ISC environments.

\subsubsection{Real Images}\label{subsub:real-images}
ISC-Perception includes real images to enhance the dataset's authenticity and variability. However, with no publicly available ISC dataset, the real images were curated from multiple sources:  still frames from ISC assembly videos, manually collected images, and publicly available datasets for humans from Roboflow Universe\cite{roboflow}.

To incorporate ISC-related real images, still frames were extracted from previous project videos. Out of 29 extracted images, 16 were extracted for annotation while removing duplicates and closely matched frames. The collected images were then annotated using Roboflow's annotation tool, followed by preprocessing steps, including brightness adjustments ($-15\%$ to $+15\%$) and rotations ($-10^\circ$ to $+10^\circ$). Additional augmentations such as $90^\circ$
 rotations, flips, and saturation adjustments generated 82 final images for the dataset.

 Furthermore, small-scale ISC members and connection plates were fabricated for manual image collection with varying appearance, position, and roation, resulting in a total of $207$ annotated images using Label Studio \cite{label_studio}. 
 
 Additionally, to address safety considerations, images of humans were included. Since numerous publicly available annotated datasets exist for humans, the Roboflow Universe dataset was utilized \cite{roboflow}. This dataset contains $235$ images of individuals in various standing and sitting poses, with preprocessing effects such as colour, brightness, shear, and stretch. All human images were manually scrutinised to address privacy concerns before inclusion in the dataset.

\section{Dataset}
\label{sec:Dataset}
ISC-Perception dataset integrates images from four primary sources as described in the section \textit{\ref{subsec:dataset-composition}}. These sources include; collected $13,399$ images via \textbf{Unity with Custom randomizers (C3)}, $3,599$ images via \textbf{SW Visualize (C1)}, $289$ images of \textbf{Real Images from previous ISC project (C4)}, and $1,728$ of \textbf{Human Images from Roboflow Universe (C5)}. The total number of images in the dataset is distributed across training, validation, and testing sets, ensuring a fair diversity and representation of different scenarios. This includes $15,974$ images of training and validation images in dataset 3 and a test set comprising $3,087$ images ($16\%$ of the total). Test images were manually selected to capture varied scenarios, ensuring robust evaluation. See Tables \ref{tab:isc_datasets_summary} and \ref{tab:dataset_distribution} for the summary and distribution of the dataset. 

\begin{table*}[htb]
\centering
\caption{\sao{Human time accounting at $N{=}\sao{10{,}000}$ images. \emph{Compute wall-clock} (GPU/CPU rendering/export) is listed in a separate column and \emph{not counted} as human labour. At our full dataset size ($N{=}\sao{15{,}974}$): manual $=\sao{266.2}$\,h, synthetic human $=\sao{30.7}$\,h, compute $\approx\sao{19.2}$\,h.}}
\label{tab:time_to_dataset}
\begin{tabular}{lccc}
\hline
\sao{\textbf{Stage}} & \sao{\textbf{Human time}} & \sao{\textbf{Compute time}} & \sao{\textbf{Notes}} \\
\hline
\sao{Unity setup \& assets}            & \sao{6.0\,h}   & \sao{--}    & \sao{project, import, materials} \\
\sao{Scene/physics authoring}          & \sao{6.0\,h}   & \sao{--}    & \sao{colliders, dynamics} \\
\sao{Sensor/exporters}                 & \sao{3.5\,h}   & \sao{--}    & \sao{RGB, depth, masks, GT} \\
\sao{Domain randomisation}             & \sao{3.5\,h}   & \sao{--}    & \sao{poses, lights, textures} \\
\sao{SolidWorks Visualize integration} & \sao{3.0\,h}   & \sao{--}    & \sao{mesh overlays, QA hooks} \\
\sao{Render/export automation}         & \sao{2.0\,h}   & \sao{12\,h} & \sao{batch scripts; GPU/CPU wall-clock} \\
\sao{QA sampling (}\sao{2\%}\sao{ @ }\sao{6\,s/image}\sao{)} & \sao{0.33\,h}  & \sao{--}    & \sao{visual checks only} \\
\sao{Final end-to-end checks}          & \sao{6.2\,h}   & \sao{--}    & \sao{splits, metadata, hashes} \\
\hline
\sao{Manual baseline (}\sao{60\,s/image}\sao{)} & \sao{\textbf{166.7\,h}} & \sao{--} & \sao{single annotator; 5-instance pilot }\sao{60\,s}\sao{ (max }\sao{80\,s}\sao{)} \\
\hline
\sao{\textbf{Synthetic total (human)}} & \sao{\textbf{30.5\,h}}  & \sao{\textbf{12\,h}} & \sao{effective }\sao{11.0\,s/image}\sao{ human time} \\
\hline
\end{tabular}
\vspace{1mm}

\noindent\footnotesize\sao{\emph{Note.} Totals in the table correspond to $N{=}\sao{10{,}000}$ (compute $\sao{=12}$\,h). 

At our full dataset size $N{=}\sao{15{,}974}$, compute is $\approx\sao{19.2}$\,h; we report this separately in the text.}
\end{table*}

\sao{\subsection{Time-to-Dataset Accounting (Synthetic vs Manual)}\label{subsec:time_to_dataset}
Table~\ref{tab:time_to_dataset} details the \emph{human} effort for our Unity-based pipeline (30.5\,h total; effective 11.0\,s/image) and lists compute wall-clock separately (12\,h). At our full dataset size ($N{=}15{,}974$), human time \footnote{Human time is dominated by fixed setup; the only $N$-dependent component is QA (2\% at 6\,s/image), which accounts for the 0.2\,h increase from 10k to 15{,}974 images. Compute wall-clock (render/export) is reported separately and not counted as labour.} is $30.5$\,h versus $266.2$\,h for manual labelling at $60$\,s/image ($-88.5\%$). Compute wall-clock (render/export) scales linearly with $N$ and is $\approx19.2$\,h; we exclude this from human-time totals.}
\sao{All stages except quality assurance (QA) sampling are fixed one-off tasks; only the QA term scales with $N$ (2\% at 6\,s/image). Hence 30.5\,h at $N{=}10{,}000$ versus 30.7\,h at $N{=}15{,}974$.}

\subsection{Statistics of the Datasets}
\begin{figure*}[!ht]
    \centering
    \includegraphics[width=7.3in]{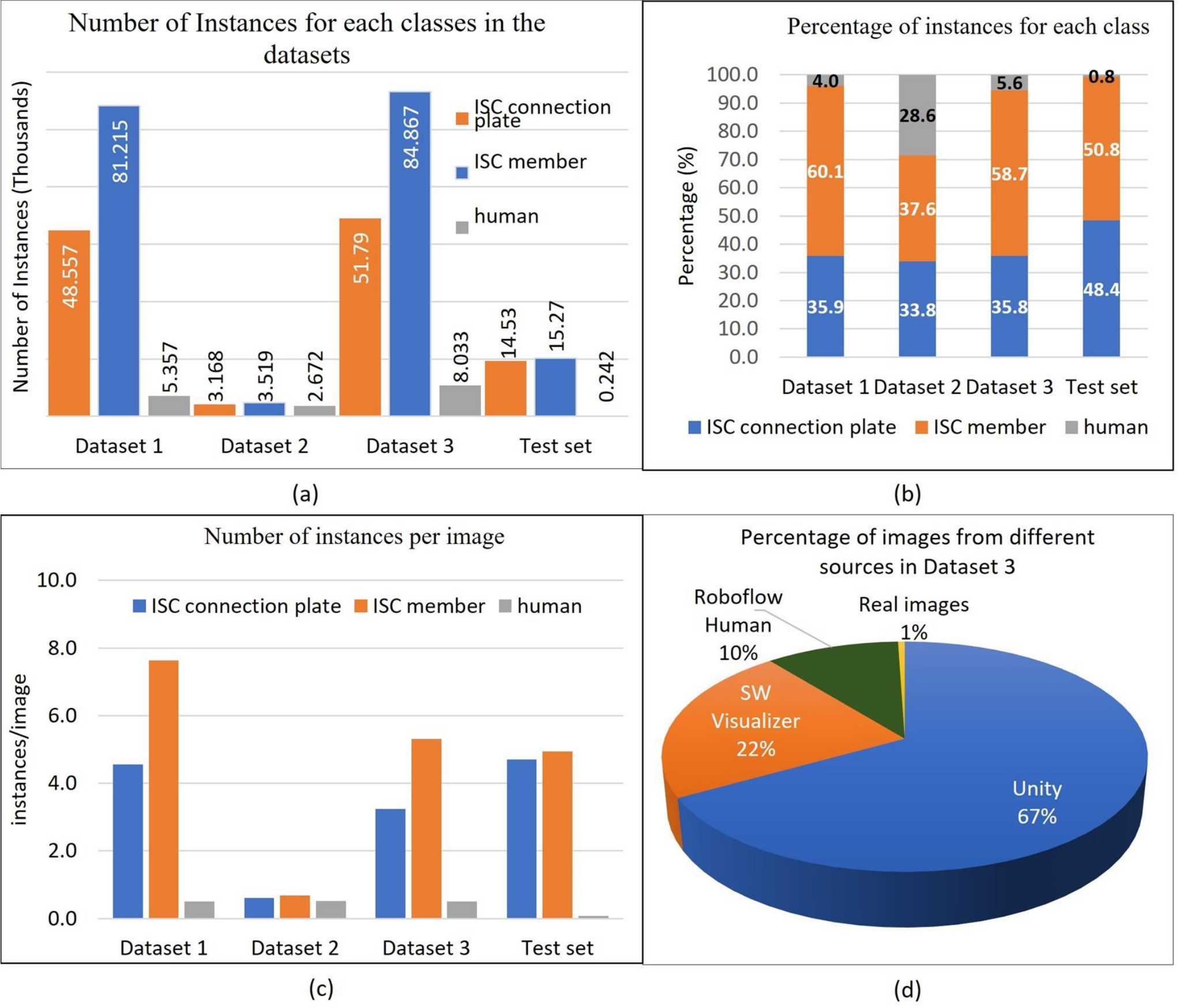}%
    \caption{Dataset statistics; (a) number of instances for each class and (b) percentage of instances in each dataset, (c)number of instances per image for each class, (d) percentage of images from different source in dataset 3.}
    \label{fig:dataset_tatistics_fig}
\end{figure*}

To evaluate the model's performance, three versions of ISC-Perception datasets were created, with a constant test set across all four versions. Table \ref{tab:dataset_distribution} provides a detailed distribution of images across datasets, while Fig.~\ref{fig:dataset_tatistics_fig} illustrates key statistics.

\begin{table}[htbp]
\caption{Summary of the Datasets for Detecting ISC Components}
\label{tab:isc_datasets_summary}
\centering
\setlength{\tabcolsep}{4pt}
\begin{tabular}{p{2.5cm}p{4.5cm}c}
\hline
\textbf{Dataset} & \textbf{Source of Images} & \textbf{Images} \\ 
\hline
Dataset 1 (Custom randomizer dataset) & Unity (custom randomizer): indoor/outdoor scenes (C3) & 10,651 \\
Dataset 2 (SW visualize and Roboflow dataset) & SW Visualize + Roboflow human annotations (C1+C5) & 5,195 \\
Dataset 3 (Hybrid dataset) & Unity + SW Visualize + Roboflow + real images (C1+C3+C4+C5) & 15,974 \\
\hline
\end{tabular}
\end{table}

\begin{table}[htbp]
\caption{Distribution of Images in ISC-Perception Across Sources}
\label{tab:dataset_distribution}
\centering
\setlength{\tabcolsep}{3pt}
\begin{tabular}{p{3.1cm}cccc}
\hline
\textbf{Source} & \textbf{Dataset 1} & \textbf{Dataset 2} & \textbf{Dataset 3} & \textbf{Test Set} \\ 
\hline
Unity (Custom randomizer) & 10,651 & 0 & 10,651 & 2,748 \\
SW Visualize & 0 & 3,551 & 3,551 & 48 \\
Roboflow Human & 0 & 1,644 & 1,644 & 84 \\
Real Images & 0 & 0 & 82 & 207 \\[2pt]
\hline
\textbf{Total} & \textbf{10,651} & \textbf{5,195} & \textbf{15,974} & \textbf{3,087} \\
\hline
\end{tabular}
\end{table}

\begin{enumerate}
    \item \textit{Image Distribution:} Dataset 3 (Hybrid Dataset) contains the largest number of images ($15,974$) from all sources, while Dataset 2 (SW Visualize with Roboflow Human) has fewer images ($5,195$). Dataset 1 (Unity Custom randomizer) focuses solely on synthetic data, with $10,651$ images.
    \item  \textit{Instances per Class}: Fig. ~\ref{fig:dataset_tatistics_fig}(a) shows that ISC members dominate with 15,270 instances in the test set, followed by ISC connection plates (14,530) and humans (242). Dataset 1 has the highest average number of instances per image for each class, as shown in Fig. ~\ref{fig:dataset_tatistics_fig}(c).
    \item \textit{Percentage of Sources}:  In Dataset 3, $67\%$ of images come from Unity, $22\%$ from SolidWorks Visualize, $10\%$ from Roboflow Universe, and $1\%$ from real ISC images (Fig. ~\ref{fig:dataset_tatistics_fig}(d)). \mrh{ISC is a newly developed novel connection that has not yet been commercially adopted in construction, so it is difficult to get real images of ISC. Hence, only 82 real images of ISC could be collected and included in the dataset, as shown in Table \protect~\ref{tab:dataset_distribution}.}
\end{enumerate}

\subsection{Example Images}\label{subsec:example-images}
\begin{figure*}[!ht]
    \centering
    \includegraphics[width=7.3in]{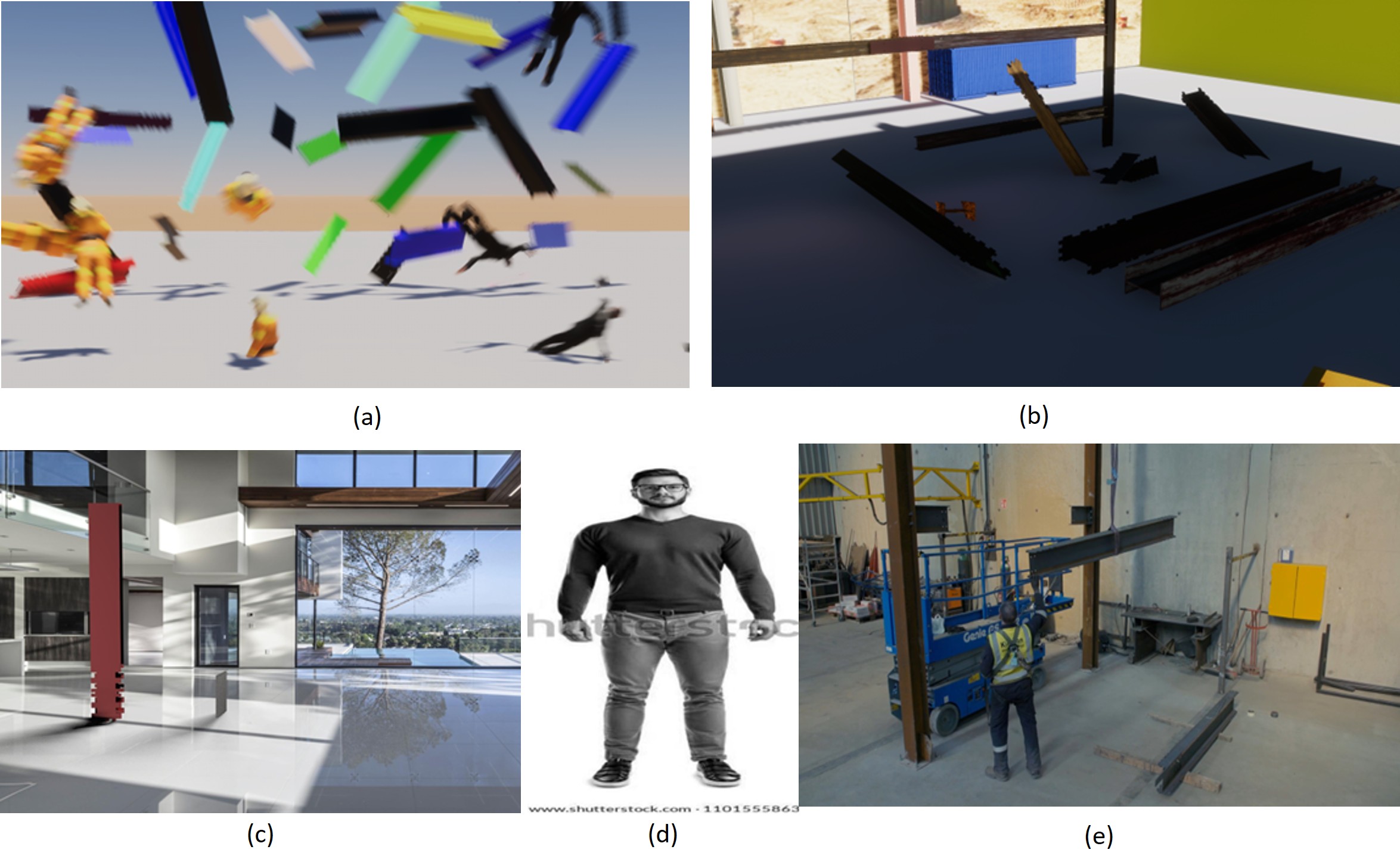}%
    \caption{\sao{Representative samples from ISC-Perception:
(a) Unity (built-in randomizers, C2),
(b) Unity (custom randomizers, C3),
(c) SolidWorks Visualize photorealistic render (C1),
(d) Human example from Roboflow Universe (C5),
(e) Real ISC frame (C4). Roboflow Universe aggregates contributions from multiple providers (which can include stock libraries); we therefore cite Roboflow Universe as the source for (d).}
    \label{fig:sample-images}}
\end{figure*}

Figure ~\ref{fig:sample-images} showcases sample images from the dataset:
\begin{enumerate}
    \item \textit{Unity Synthetic Images:} Fig. \ref{fig:sample-images}(a) and Fig. \ref{fig:sample-images}(b) highlight images generated using built-in and custom randomizers, with varying lighting, object placements, and occlusion effects.
    \item \mrh{\textit{Photorealistic Images:}} Fig. \ref{fig:sample-images}(c) illustrates high-quality images from SolidWorks Visualize, featuring diverse scenes and objects, including grayscale and construction site settings.
    \item \textit{Real Images:} Fig. \ref{fig:sample-images}(d) shows human images from Roboflow Universe, while Fig. \ref{fig:sample-images}(e) presents real images from previous ISC assembly projects. \sao{Roboflow Universe aggregates community-contributed images from multiple providers (which may include stock-photography sources); we therefore cite the dataset as Roboflow Universe for panel (d).}
\end{enumerate}

\section{Performance Analysis}
\label{sec:Performance-Analysis}
Performance analysis of the datasets was divided into two categories. Initially, three different computer vision models were generated by training the YOLOv8 algorithm with three different datasets (dataset 1, dataset 2 and dataset 3) from Table \ref{tab:dataset_distribution}. The training performance was initially analysed using several performance metrics. Finally, the trained model was applied to the test set from Table \ref{tab:dataset_distribution} to predict the desired object. The effect of different image types and datasets were analysed based on the prediction performance. \sao{We first report full-size training results, then a controlled size-matched comparison to disentangle composition from dataset size (Sec.~\ref{subsec:size_matched}).}

\subsection{Hardware configuration}\label{subsec:implementation-details} 
To create the object detection model, the YOLOv8 algorithm was trained with all three versions of the dataset. An Alienware m16 laptop configured with core i9 13900HX processor, 32.0GB RAM and Nvidia GeForce RTX 4060 12GB GDDR6 graphics card was used to train the ISC components detection model. Each dataset was trained for a maximum 250 epochs, with 30 epochs patience as a stopping criterion. Other training parameters remained at default. The best model was saved for each version of the dataset.

\subsection{Training settings}\label{subsec:train-settings}
\sao{We trained YOLOv8n (Ultralytics v8.3.198) starting from pretrained model on COCO \cite{jocher_yolo_2023} using the Ultralytics trainer. We used an input size 640$\times$640, batch size 16, and random seed 42. We enabled a cosine learning-rate schedule and do not override the framework’s base learning rate; other optimizer hyperparameters remain at defaults. For the main experiments, we used early stopping with a cap of 250 epochs and patience 30, selecting the best checkpoint by validation mAP. For the controlled size-matched comparison (Section.~\ref{subsec:size_matched}), we fixed the number of optimizer updates (no early stopping) and match augmentations, input size, batch size, and learning rate schedule across conditions.}

\subsection{Testing procedure}\label{subsec:testing_procedure} 
The best models trained were obtained from a collective of three distinct dataset configurations: the hybrid dataset (which includes custom randomizer, SW \mrh{Visualize}, Roboflow, and manually annotated images), a dataset using only the custom randomizer images, and a dataset using only the SW \mrh{Visualize} images. Tests were then conducted using two groups of test data:
\begin{enumerate}
    \item Complete test set: This contains samples from all sources – custom randomizer, SW Visualize, Roboflow Human, and manually annotated images. This contains a total of 3087 images. Details of the number of samples from each source are presented in Table \ref{tab:dataset_distribution}.
    \item Small scale bench test: This set comprises real-world samples using a multi-view, small scale experimental setup of robotic assembly of ISC, where synchronised cameras provided multiple perspectives of the same scene. This setup allows us to do continuous object detection and tracking of ISC components and humans.
\end{enumerate}
\sao{To disentangle composition from dataset size, we also evaluated models trained on size- and class-matched subsets; see Section~\ref{subsec:size_matched}.}

\subsection{Results and Discussion: Testing with test set}\label{sec:results-and-discussion}
\subsubsection{Results Overview}\label{subsub:Results-overview}
The three models trained on Hybrid dataset, Custom randomizer dataset, and SW Visualize and Roboflow dataset – were all tested on a combined test set (Table \ref{tab:dataset_distribution}) containing 3,087 images- includes samples from all data sources; Unity custom randomizer, SW \mrh{Visualize}, Roboflow human images, and manually annotated real-world images.
Four performance \sao{metrics} ( Precision, Recall, mAP@0.5 and mAP@[0.5:0.95]) were used to assess the trained performance of the model across the test data. We present the results of training in Table \ref{tab:performance_metrics}.

The model trained on the Hybrid dataset achieved the highest overall performance with a mAP (50-95) of 0.664 compared to 0.564 for custom randomizer dataset and 0.321 for SW Visualize and Roboflow dataset indicating that it is able to generalise across a diverse range of image types (see Table ~\ref{tab:performance_metrics}). The performance was particularly strong for human detection, where it achieved a mAP@[0.5:0.95] of 0.804 and high precision and recall scores. On the other hand, the model’s performance for ISC connection plates was lower with a mAP@[0.5:0.95] of 0.523 likely due to the complexity of detecting these components in varied real-world environments (Table \ref{tab:performance_metrics}).

\subsubsection{Controlled Size-Matched Comparison}\label{subsec:size_matched}
\sao{We trained the model with identical hyperparameters on size- and class-matched subsets ($N{=}5{,}195$) of \emph{Dataset-1}, \emph{Dataset-2}, and the \emph{Dataset-3}. Subsets are constructed by stratified sampling to preserve class priors (and, where available, instances-per-image bins); the test set is fixed (3{,}087 images). We equalised the training budget by using a fixed number of optimiser updates (no early stopping), and we match augmentations, input resolution, batch size, and learning-rate schedule across conditions as in the main training experiment. At fixed $N$, the hybrid composition achieves mAP@0.50 of $0.675$ and mAP@[0.50:0.95] $0.549$, exceeding Dataset-1 ($0.546/0.430$) and Dataset-2 ($0.249/0.206$), with higher precision (0.830 vs 0.776/0.649) and recall (0.625 vs 0.505/0.146). This corresponds to $+0.129$ mAP@0.50 ($+23.6\%$) and $+0.119$ mAP@[.50:.95] ($+27.7\%$) over Dataset-1, and $+0.426$ / $+0.343$ ($+171\%$ / $+166\%$) over Dataset-2. These gains at constant size indicate the improvement stems from the \emph{hybrid composition}, not merely dataset size. See Table~\ref{tab:size_matched_5195}.}

\begin{table}[!t]
\centering
\footnotesize                      
\setlength{\tabcolsep}{2pt}        
\caption{Performance of trained models on the test set}
\label{tab:performance_metrics}
\resizebox{\columnwidth}{!}{       
\begin{tabular}{|l|c|c|c|c|c|}
\hline
\multicolumn{1}{l|}{}                 & \textbf{Dataset} & \textbf{precision} & \textbf{recall} & \textbf{mAP@0.5} & \textbf{mAP@[0.5:0.95]} \\ \hline
\multirow{3}{*}{\textbf{Overall}}              & 1       & 0.82      & 0.52   & 0.66    & 0.56       \\ \cline{2-6} 
                                      & 2       & 0.40      & 0.32   & 0.39    & 0.32       \\ \cline{2-6} 
                                      & 3       & 0.85      & 0.67   & 0.75    & 0.66       \\ \hline
\multirow{3}{*}{\textbf{ISC connection plate}} & 1       & 0.83      & 0.47   & 0.64    & 0.53       \\ \cline{2-6} 
                                      & 2       & 0.36      & 0.21   & 0.21    & 0.18       \\ \cline{2-6} 
                                      & 3       & 0.81      & 0.49   & 0.64    & 0.52       \\ \hline
\multirow{3}{*}{\textbf{ISC member}}           & 1       & 0.80      & 0.71   & 0.75    & 0.67       \\ \cline{2-6} 
                                      & 2       & 0.52      & 0.16   & 0.33    & 0.24       \\ \cline{2-6} 
                                      & 3       & 0.80      & 0.73   & 0.76    & 0.66       \\ \hline
\multirow{3}{*}{\textbf{Human}}                & 1       & 0.82      & 0.38   & 0.59    & 0.50       \\ \cline{2-6} 
                                      & 2       & 0.33      & 0.67   & 0.61    & 0.54       \\ \cline{2-6} 
                                      & 3       & 0.92      & 0.78   & 0.87    & 0.80       \\ \hline
\end{tabular}}
\end{table}

\begin{table}[t]
\centering
\caption{\sao{Size-matched comparison. Dataset-2 uses its full train set ($N{=}5{,}195$); Dataset-1 and Dataset-3 are stratified-sampled to $N{=}5{,}195$. Test set fixed (3{,}087). $^\ast$ sampled; $^\dagger$ full.}}
\label{tab:size_matched_5195}

\small                                 
\setlength{\tabcolsep}{3.5pt}          
\renewcommand{\arraystretch}{0.98}     

\begin{tabular}{@{}lcccc@{}}
\hline
\sao{Train set ($N$)} & \sao{mAP50} & \sao{mAP50--95} & \sao{Prec.} & \sao{Rec.} \\
\hline
\sao{Dataset-1$^\ast$ (5{,}195)}                    & \sao{0.546} & \sao{0.430} & \sao{0.776} & \sao{0.505} \\
\sao{Dataset-2$^\dagger$ (5{,}195)}                 & \sao{0.249} & \sao{0.206} & \sao{0.649} & \sao{0.146} \\
\sao{\textbf{Dataset-3$^\ast$ (hybrid, 5{,}195)}}   & \sao{\textbf{0.675}} & \sao{\textbf{0.549}} & \sao{\textbf{0.830}} & \sao{\textbf{0.625}} \\
\hline
\end{tabular}
\end{table}

\subsubsection{Confusion Matrix and Performance Curves}\label{subsubsec:confusion-matrices}
\begin{figure*}[!ht]
    \centering
    \includegraphics[width=7in]{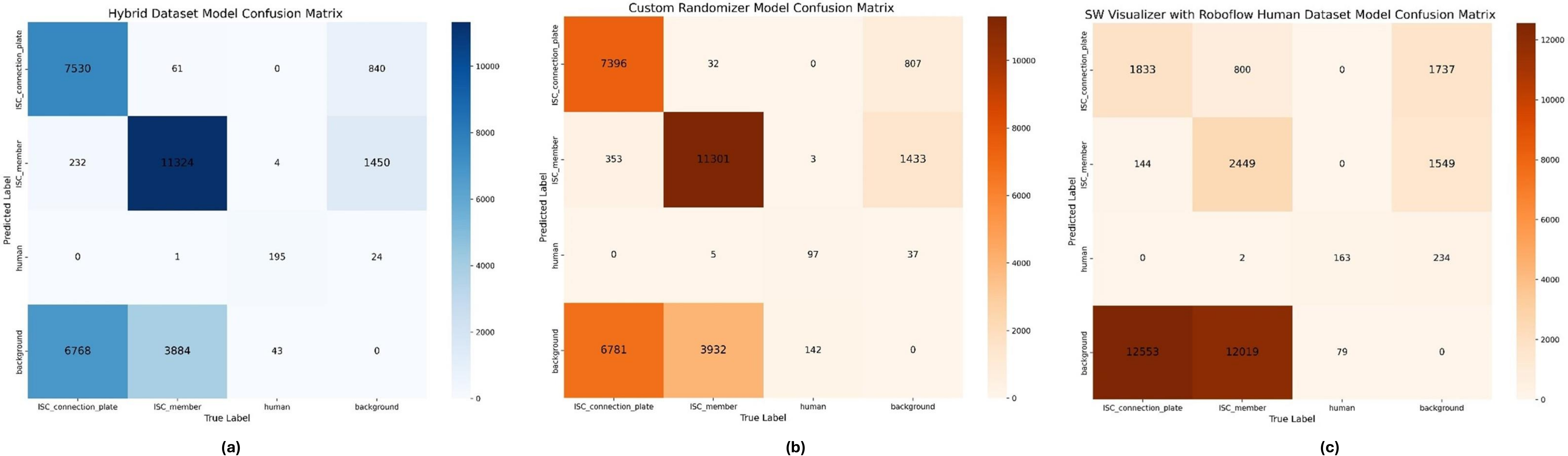}%
    \caption{Confusion Matrix plots of trained models on test Set; (a) on Hybrid Dataset (b) Custom Randomizer Dataset (c) SW Visualize with Roboflow Dataset}
    \label{fig:Confusion-Matrix-plots}
\end{figure*}

\begin{figure*}[!ht]
    \centering
    \includegraphics[width=5.3in]{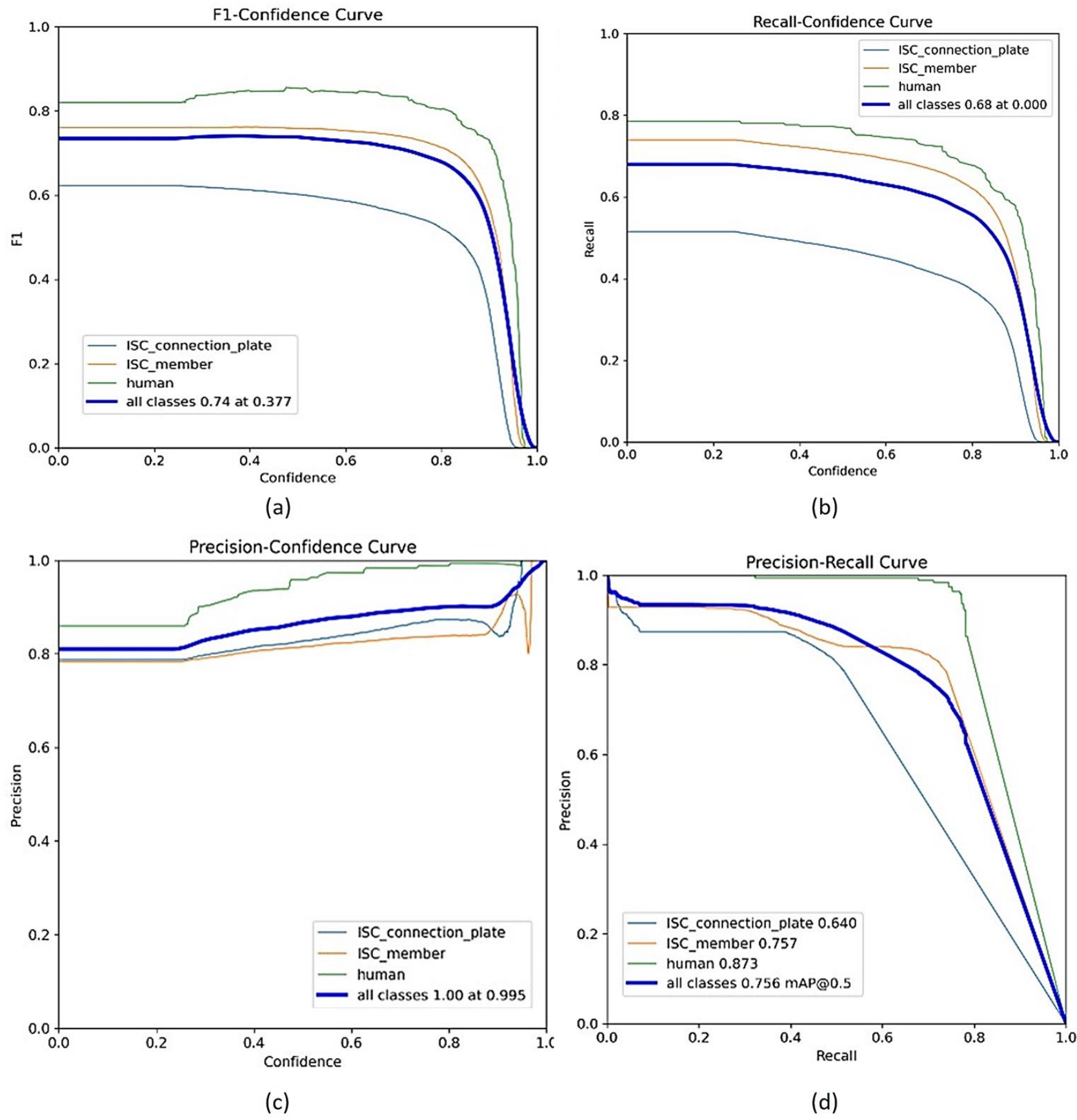}%
    \caption{Performance curves of the model trained on Hybrid Dataset Evaluated on Test Set; (a) F1 confidence curve, (b) recall-confidence curve, (c) precision-confidence curve, (d) precision-recall curve.}
    \label{fig:Performance-curves}
\end{figure*}

As shown in the confusion matrix in Fig.~\ref{fig:Confusion-Matrix-plots} the model trained on hybrid dataset correctly identified a large proportion of ISC components and human instances compared to model trained on custom randomizer dataset and SW Visualize and Roboflow human dataset. However, all trained model exhibited some misclassification. For example, the trained model successfully detected 7530 connection plates while there are only 840 false negatives for connection plates and the background and only 61 false negatives for connection plates and the ISC members (Fig.\ref{fig:Confusion-Matrix-plots}(a)). 
Model trained on the hybrid dataset also correctly detected $11,324$ instances of ISC members and 195 instances of human (Fig.\ref{fig:Confusion-Matrix-plots}(a)). However, as per Fig.\ref{fig:Confusion-Matrix-plots}(c), model trained on the SW Visualize and Roboflow human performed worst by correctly detecting only $1,883$ instances of ISC connection plate, $2,449$ instances of ISC member and 163 instances of human on the test set. As per the Fig.\ref{fig:Performance-curves}(a), F1-Confidence Curve showing an overall F1 score of 0.74 at a confidence threshold of 0.377. Precision remained high for all classes, as demonstrated in the Precision-Confidence Curve  Fig.\ref{fig:Performance-curves}(c), although ISC connection plate detection showed a noticeable drop-off in recall confirming that the model sometimes missed these components in challenging scenes.

\subsubsection{Discussion on Model Performance Comparison: Hybrid dataset vs. Custom randomizer dataset vs. SW Visualize and Roboflow dataset}\label{subsubsec:model-performance}
The evaluation results of the model trained on the three distinct datasets are presented in Table \ref{tab:performance_metrics}, which summarise their performance metrics for each class.

\textit{Hybrid dataset model:}
The model trained on the Hybrid dataset (Table \ref{tab:performance_metrics}) demonstrated the highest overall performance across all object classes. It achieved a Box Precision of 0.846 and a Recall of 0.666, leading to an overall mAP@0.5 of 0.756 and a mAP@[0.5:0.95] of 0.664. The performance in detecting ISC connection plates was slightly lower, with a mAP@[0.5:0.95] of 0.523, indicating some difficulty in precise identification. However, the model excelled in ISC member detection, attaining a mAP@[0.5:0.95] of 0.664, and performed exceptionally well in detecting humans, with a mAP@[0.5:0.95] of 0.804 and a Recall of 0.777.

\textit{Custom randomizer dataset model:}
The Custom randomizer dataset model (Table \ref{tab:performance_metrics}) displayed a similar trend, but with a lower overall performance compared to the Hybrid dataset model. The Box Precision of 0.818 and Recall of 0.521 resulted in an overall mAP@0.5 of 0.659 and mAP@[0.5:0.95] of 0.564. For ISC connection plates, the performance was comparable to the Hybrid dataset model with a mAP@[0.5:0.95] of 0.523. However, the model exhibited reduced accuracy in detecting humans, with a mAP@[0.5:0.95] of 0.503, indicating limitations in handling more diverse human instances.

\textit{SW Visualize and Roboflow dataset model:}
The model trained on the SW Visualize and Roboflow dataset (Table \ref{tab:performance_metrics}) performed the weakest overall, reflecting its narrow focus on the data set. It achieved a much lower Box Precision of 0.404 and Recall of 0.321, resulting in an overall mAP@0.5 of 0.386 and mAP@[0.5:0.95] of 0.321. For ISC connection plates, the mAP@[0.5:0.95] was the lowest at 0.176, and ISC member detection also lagged, with a mAP@[0.5:0.95] of 0.244. While this model was relatively better at human detection with a mAP@[0.5:0.95] of 0.541, its overall ability to generalise to ISC components was clearly limited. From these results, it is evident that the Hybrid Dataset model provides the best performance across all object categories, particularly excelling in human detection and ISC member identification. The Custom Randomizer model, while decent, struggles with human detection and generalisation to real-world data. Lastly, the SW Visualize and Roboflow Human model shows clear limitations, particularly for ISC components, due to its narrow training focus.

\subsection{Small scale bench testing}\label{subsec:testing-in-the-wild}
\begin{figure*}[!ht]
    \centering
    \includegraphics[width=7in]{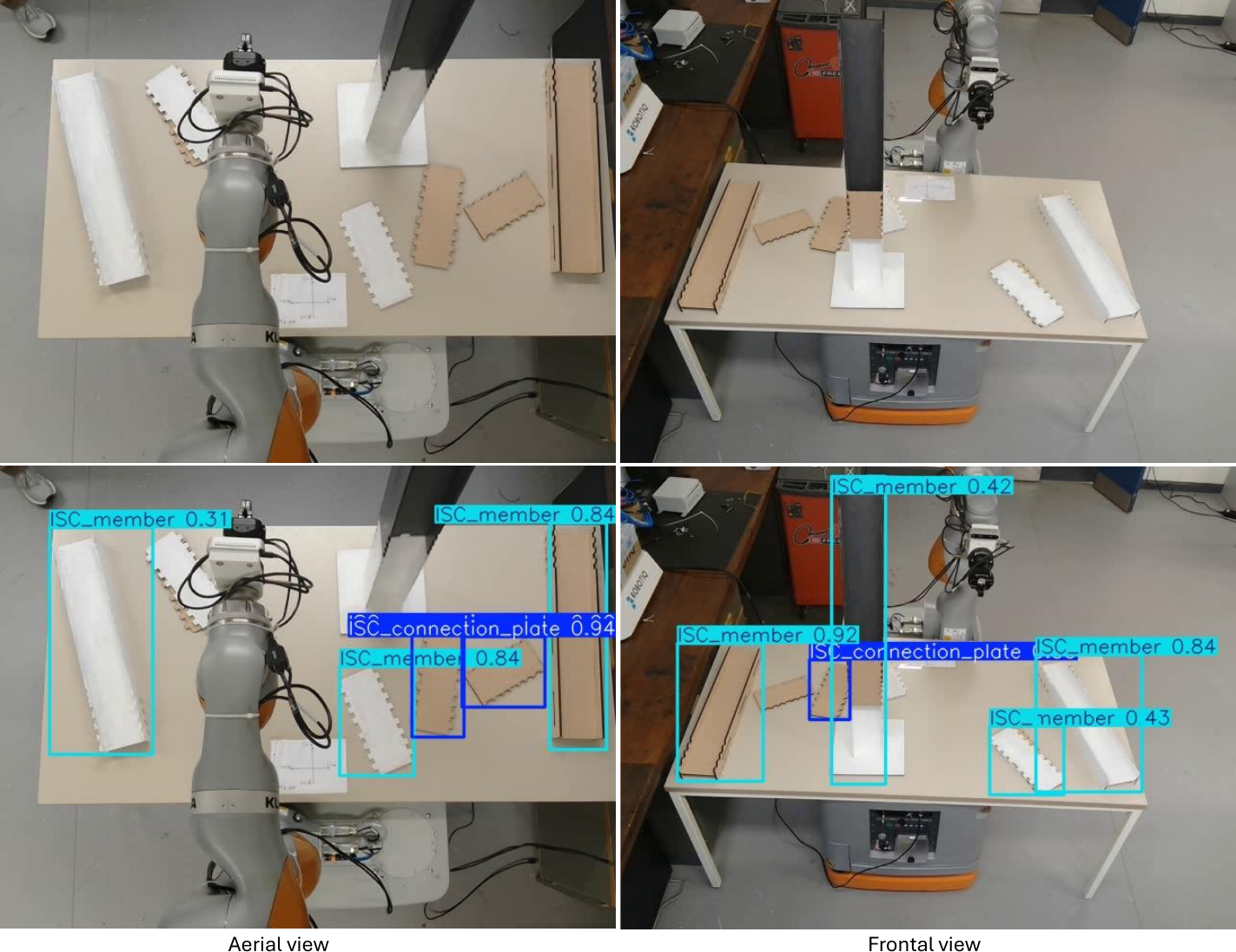}%
    \caption{Synchronized aerial- and frontal-camera views of the bench-top ISC assembly experiment, shown before (top row) and after (bottom row) YOLOv8 inference.}
    \label{fig:testing-in-the-wild}
\end{figure*}

\begin{figure*}[!ht]
    \centering
    \includegraphics[width=7in]{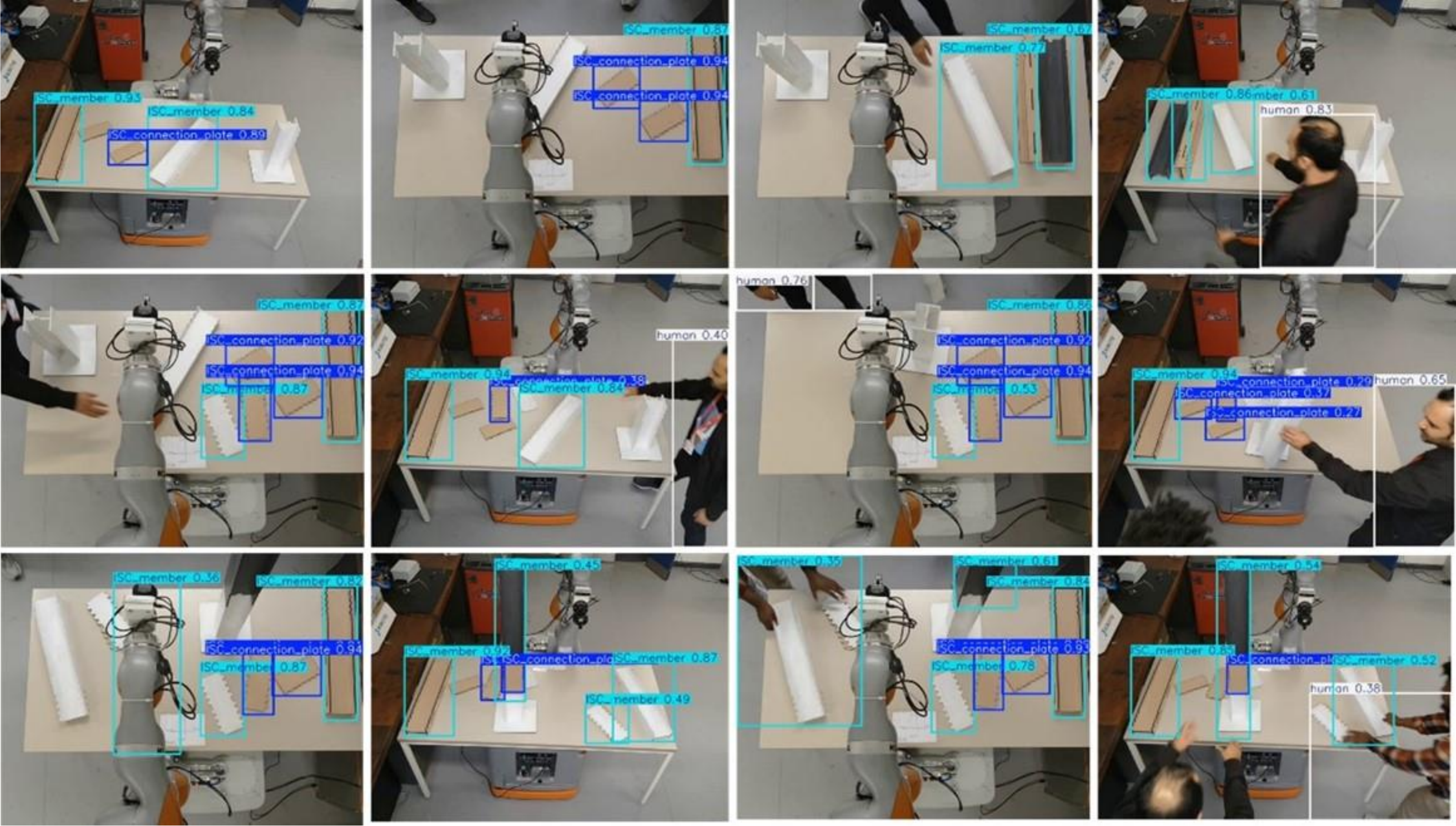}%
    \caption{Real-Time Object Detection Tracking Performance on ISC Objects, Connection Plates, and Human Workers.}
    \label{fig:real-time-object-tracking}
\end{figure*}

\begin{figure*}[!ht]
    \centering
    \includegraphics[width=7in]{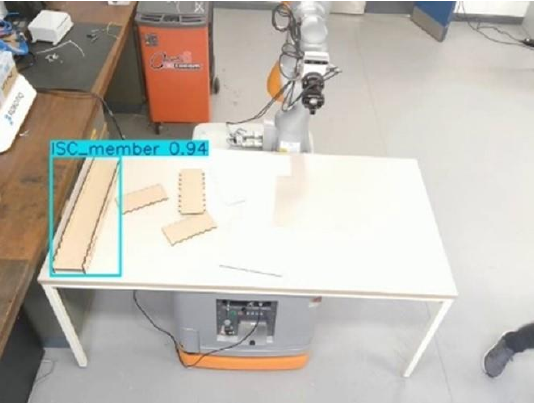}%
    \caption{Glaring in the frontal view impacts detection performance.}
    \label{fig:glaring-frontal-view}
\end{figure*}

The overall objective of our research is to develop and integrate the object detection model into a broader robotic assembly process. To assess the robustness and performance of the model in real-world conditions, we incorporated its output into the vision module of our assembly framework. The object detection model serves as an intermediary for subsequent vision-based tasks within this framework. Our setup, as shown in Fig. \ref{fig:testing-in-the-wild} consists of a multi-camera system, where synchronised views provide multiple perspectives to minimise occlusion and enhance overall detection robustness. \sao{From a 2\,min, 60\,fps bench-test video with two synchronised side views (Fig.~\ref{fig:testing-in-the-wild}), we temporally subsampled every 10th frame to reduce correlation and manually annotated the resulting $\sim$1{,}200 frames for ISC components and humans. Using standard detection metrics, the detector achieved mAP@0.50 $=0.943$, mAP@[.50:.95] $=0.823$, precision $=0.951$, and recall $=0.930$.} 
 (See Fig. \ref{fig:real-time-object-tracking} for detailed detection result samples). However, the system was not without its challenges. \sao{Failure cases were primarily due to glare in the front-facing camera view (Fig.~\ref{fig:glaring-frontal-view}), leading to occasional missed detections of connection plates; improving robustness to challenging lighting and appearance shifts is left for future work.}

\section{Conclusions}\label{sec:conclusions}
This research demonstrated a procedural approach to generate a hybrid dataset for detecting ISC components in a steel-structure assembly site. As the collection of real images from the steel-structure assembly site is an arduous and unsafe task, synthetic and photorealistic images were created to compensate for the need for real images. Synthetic images from Unity 3D provide annotated images, while \mrh{photorealistic images from SW Visualize} require manual annotation. Multiple datasets were created using various types of images. Dataset 1 was created using only synthetic images generated with Unity's custom randomizers. Dataset 2 was created using photorealistic images from SW \mrh{Visualize} and real human images from the Roboflow Universe public dataset. Dataset 3 was created from Unity's custom randomizer, SW \mrh{Visualize} and Roboflow, and real-world images. Only 3 object classes (ISC connection plate, ISC member, and human) were selected for object detection tasks. These were selected because the robot will only manipulate the ISC connection plate and ISC member, while the human was added to ensure the safety.

Testing results showed that models trained on the hybrid dataset (Dataset 3) outperformed those trained on either synthetic (Dataset 1) or photorealistic data alone (Dataset 2). The hybrid dataset model demonstrated superior precision and recall across all object classes (ISC connection plate, ISC member, and human) when tested on the complete test set. The custom randomizer (Dataset 1) model achieved reasonable performance in testing but still lagged behind the hybrid model. The model trained on SW Visualize and Roboflow Human images (Dataset 2) had the lowest performance, especially in detecting ISC connection plates and ISC members, highlighting the difficulty of generalising from such a limited dataset.

To improve the impact of the synthetic images in the hybrid dataset, better quality and realistic simulation scenarios will be created in the future work. With proper computer aided design tools and graphics software, accurate colour, shape and scale will be created for the objects in the steel structure assembly site scene in Unity. Moreover, the manual annotation of the photorealistic images will be converted to an automatic annotation process to save time and create more variations. More natural lighting and other environmental effects will be introduced for both Unity and the SW visualize scene to enhance the photorealism. While this process of generating a hybrid dataset is not completely automatic, this method provides a procedure to generate a hybrid dataset where the collection of real images is very difficult, and the 3D model of the target object is readily available. 
Overall, the results of this work reinforce the importance of using datasets for robust object detection in real-world industrial settings and provide a foundation for future research in automating complex tasks in the construction and assembly industries.  This method provides a scalable approach that can further be adapted to other robotic applications, particularly in challenging environments where human access is seen as restricted such as nuclear sites, tunnels, or remote workspaces.  In addition, it has been robustly shown that hybrid datasets can provide a rich way to train computer vision models especially for emerging applications were limited relevant public datasets are available. 

\section*{Acknowledgment}
The authors gratefully acknowledge the financial support provided by the National Science Foundation (NSF) award 2222815, the Science Foundation Ireland award 21/US/3797, and the Department for the Economy (DfE), UK, through grant USI-218.
\bibliography{main}

\begin{IEEEbiography}[
{\includegraphics[width=1in,height=1.25in,clip,keepaspectratio]{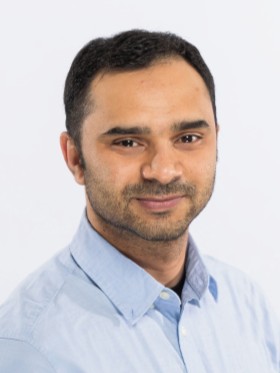}}
]{Miftahur Rahman}
is a Lecturer in Robotics at Kingston University London. His research focuses on autonomous robotic systems, mobile manipulators, sensor fusion, construction automation, and computer vision. He holds a Ph.D. in Manufacturing from Cranfield University and has contributed to several interdisciplinary projects and peer-reviewed publications in intelligent inspection and repair robotics.
\end{IEEEbiography}

\vspace{-13.5mm}
\begin{IEEEbiography}[
{\includegraphics[width=1in,height=1.25in,clip,keepaspectratio]{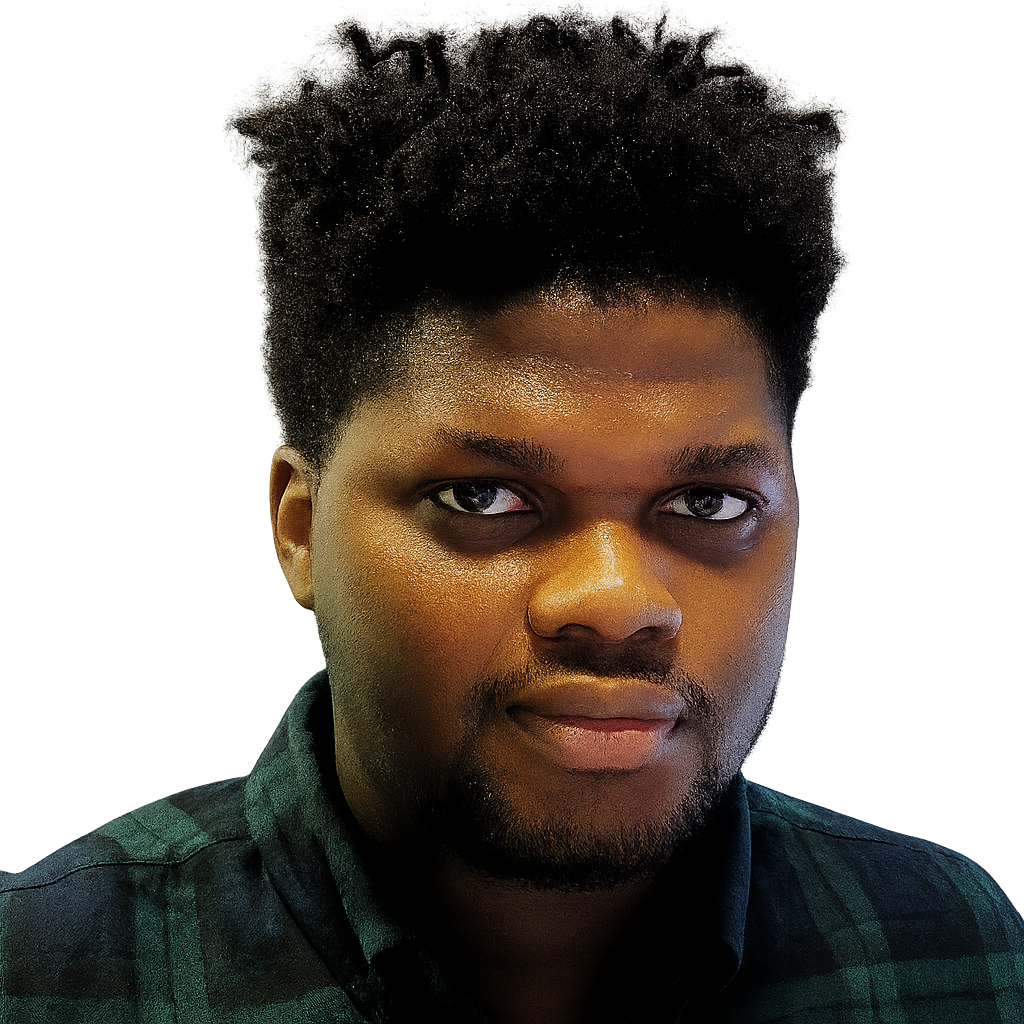}}
]{Samuel Adebayo}
(Member, IEEE) received the Ph.D.\ degree in Machine Learning from Queen’s University Belfast, U.K., in 2024. He is a Research Fellow in Computer Vision at Queen’s University Belfast. His research spans deep learning, causal machine learning, conformal prediction and the integration of psychological principles; perception, cognition, and emotion into computational intelligence. 
\end{IEEEbiography}
\vspace{-10.5mm}
\begin{IEEEbiography}[
{\includegraphics[width=1in,height=1.25in,clip,keepaspectratio]{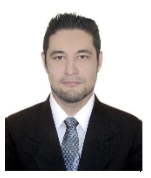}}]{Dorian A Acevedo-Mej\'ia}  is a Ph.D. Candidate in Civil and Environmental Engineering at The University of Texas at San Antonio. His research focuses on the development of self-centering horizontal structural systems to enhance the seismic performance of steel structures. He has published in Q1 journals, including the Journal of Constructional Steel Research, and in high-impact international conferences such as the World Conference on Earthquake Engineering and the World Conference on Seismic Isolation. He has over 18 years of professional experience as a structural engineer, working on infrastructure and mining projects around the world. His research interests include earthquake engineering, structural dynamics, and the design of resilient steel systems. 
\end{IEEEbiography}
\vspace{-10.5mm}
\begin{IEEEbiography}[
{\includegraphics[width=1in,height=1.25in,clip,keepaspectratio]{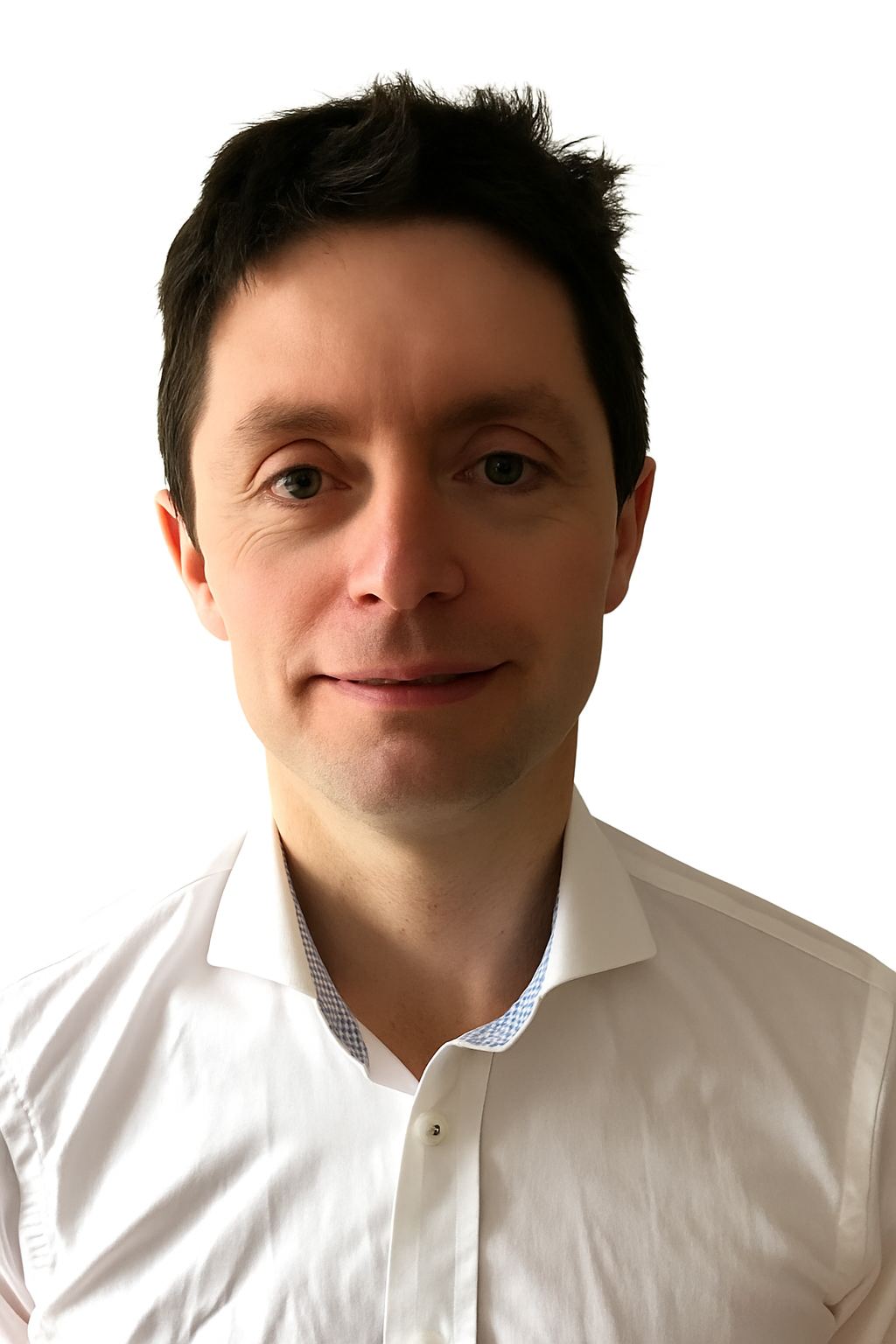}}]{David Hester}
received the B.Eng.\ degree in Civil Engineering and the Ph.D.\ degree in Bridge Structural Health Monitoring from University College Dublin, in 2000 and 2012, respectively. He is currently a Senior Lecturer in Structural Engineering with Queen’s University Belfast. His main research interests include structural dynamics and bridge structural health monitoring.
\end{IEEEbiography}
\vspace{-10.5mm}
\begin{IEEEbiography}[
{\includegraphics[width=1in,height=1.25in,clip,keepaspectratio]{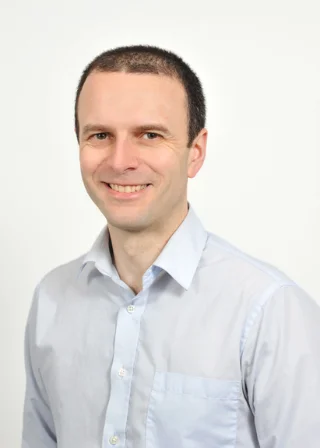}}]{Daniel McPolin}
received the Ph.D.\ degree in Structural Engineering from Queen’s University Belfast.  
He is a Senior Lecturer in the School of Natural and Built Environment at Queen’s, researching engineered-timber composites, high-performance cementitious materials and immersive digital technologies for construction.  Dr McPolin gained international recognition for the Guinness-record “world’s largest Meccano bridge’’ outreach project and is a Co-Investigator on the ARISE steel-assembly robotics programme.
\end{IEEEbiography}

\vspace{-10.5mm}
\begin{IEEEbiography}[
{\includegraphics[width=1in,height=1.25in,clip,keepaspectratio]{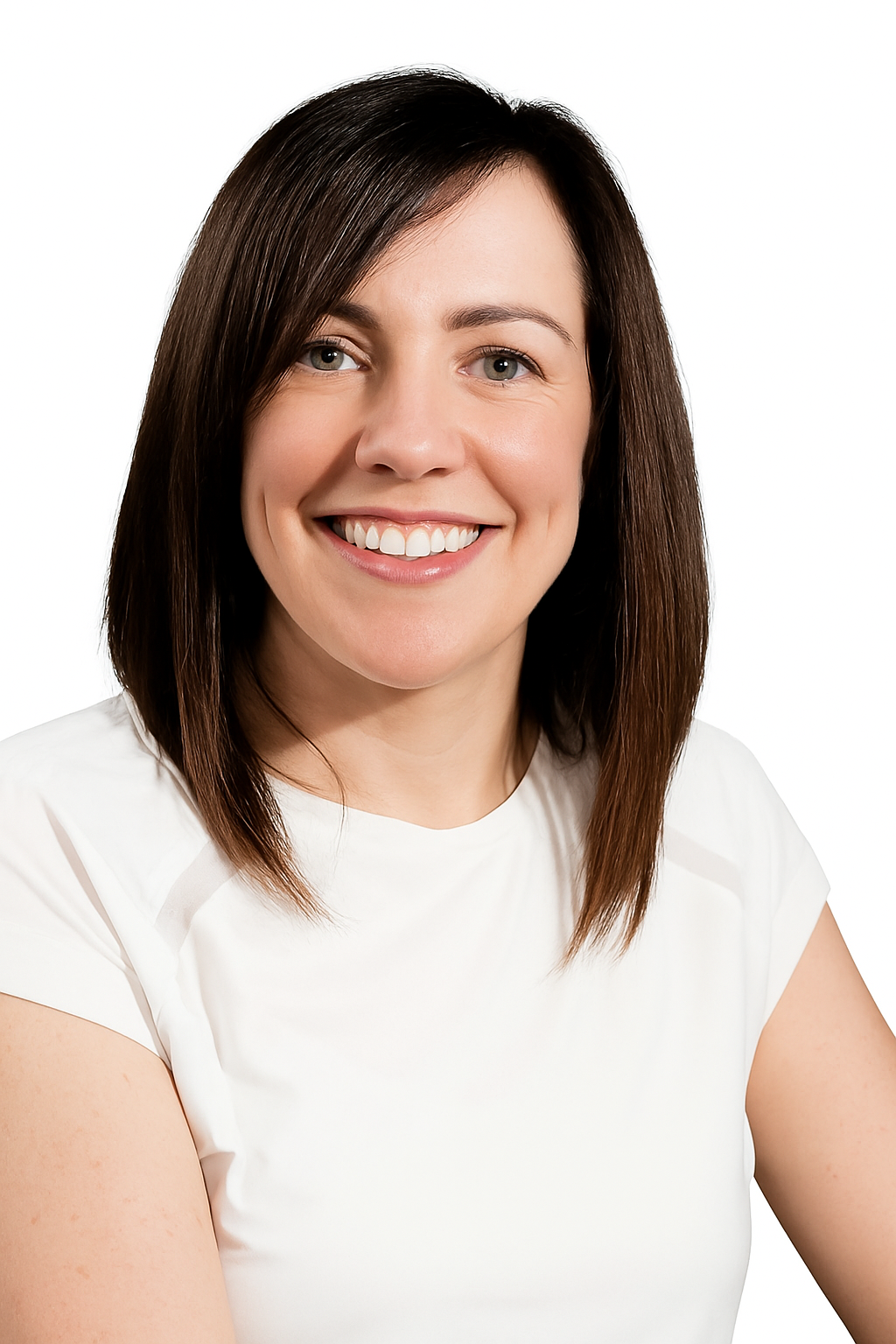}}]
{Karen Rafferty}
is currently the Head of the School of Electronics, Electrical Engineering and Computer Science, Queen’s University Belfast.  She has over fifteen years’ experience working within the fields of software engineering, sensor fusion, and real-time software development, and over ten years’ experience within the areas of virtual and augmented reality and multi-sensorial systems.  Her research interests include the application of tools and technologies to lead new disruptive practices and systems for many application areas, with a main focus on Health and Training, and Industry and Automation.
\end{IEEEbiography}
\vspace{-10.5mm}

\begin{IEEEbiography}[
{\includegraphics[width=1in,height=1.25in,clip,keepaspectratio]{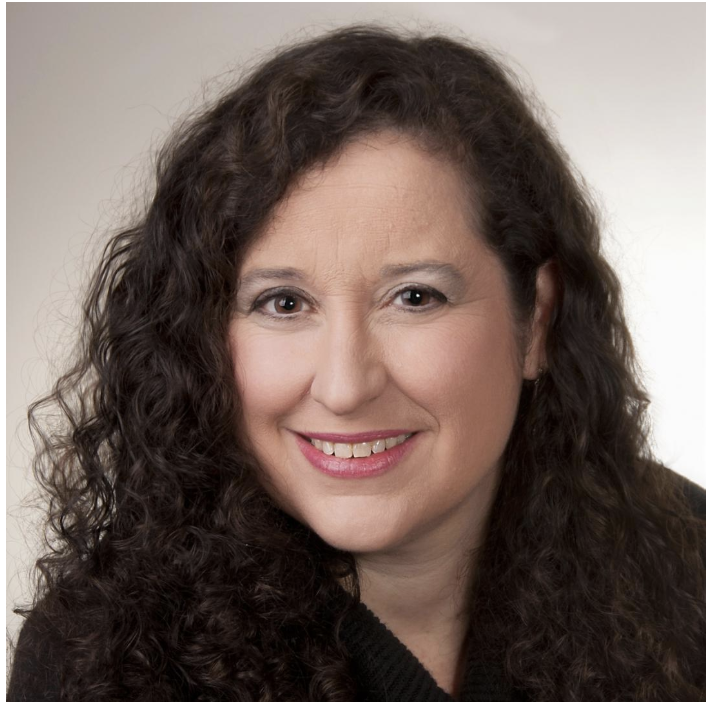}}
]{Debra F Laefer} received her Ph.D. degree in Civil Engineering from the University of Illinois Urbana-Champaign in 2001. She is a Full Professor of Urban Informatics in NYU’s Department of Civil and Urban Engineering and Center for Urban Science + Progress. Prof. Laefer’s work bridges geotechnical engineering, remote sensing and urban informatics, emphasising protection of historic fabric during subterranean construction and ultra-dense aerial LiDAR for city-scale modelling. She has published over 160 papers and holds 4 patents. She is co-inventor of the ISC with Dr. Salam Al-Sabah
\end{IEEEbiography}
\end{document}